\pdfoutput=1

\documentclass[letterpaper, 10 pt, journal, twoside]{IEEEtran}

\usepackage[T1]{fontenc}
\usepackage{mathrsfs}
\usepackage{amssymb}
\usepackage{subcaption}  

\usepackage{amsmath}
\usepackage{epsfig}
\usepackage{color}

\usepackage{hyperref}

\usepackage{amsmath, amssymb}
\usepackage{tikz}
\usepackage{graphicx}
\usepackage{geometry}
\usepackage{booktabs}
\usepackage{tabularx}  
\usepackage{algorithm}
\usepackage{algpseudocode}
\usepackage{algorithm}
\usepackage{algpseudocode}

\title{A  Consensus-Bayesian Framework for Detecting Malicious Activity in Enterprise Directory Access Graphs }

\author{Pratyush Uppuluri, Shilpa Noushad \& Sajan Kumar  \\
Purdue University \\
West Lafayette, IN, USA \\
\texttt{\{puppulur,snoushad,kumar836\}@purdue.edu}
}

\begin{document}

\maketitle

\begin{abstract}
This work presents a consensus-based Bayesian framework to detect malicious user behavior in enterprise directory access graphs. By modeling directories as topics and users as agents within a multi-level interaction graph, we simulate access evolution using influence-weighted opinion dynamics. Logical dependencies between users are encoded in dynamic matrices \( C_i \), and directory similarity is captured via a shared influence matrix \( W \). Malicious behavior is injected as cross-component logical perturbations that violate structural norms of strongly connected components (SCCs). We apply theoretical guarantees from opinion dynamics literature to determine topic convergence and detect anomaly via scaled opinion variance. To quantify uncertainty, we introduce a Bayesian anomaly scoring mechanism that evolves over time, using both static and online priors. Simulations over synthetic access graphs validate our method, demonstrating its sensitivity to logical inconsistencies and robustness under dynamic perturbation.
\end{abstract}

\section{Introduction}

Security threats in both on-premise and cloud infrastructures have become increasingly sophisticated, ranging from ransomware and malware to insider threats and unintentional access breaches~\cite{vectra2023threatdetection}. To address this, it is critical to adopt a \textbf{proactive security approach} rather than relying solely on \textbf{post-incident forensics}.Although existing system-level solutions such as \textbf{endpoint protection}, \textbf{data encryption}, and \textbf{user behavior analytics (UEBA)} offer some protection, encryption imposes performance overheads, especially for data in use. UEBA~\cite{microsoft-ueba}, which focuses on analyzing user-entity access behavior, is particularly suitable for detecting insider threats and behavioral anomalies.

In enterprise environments, users typically operate in logical groups and interact with related entities such as shared directories, repositories, or cloud services. For instance, a team of engineers working on a common project may frequently access a set of GitHub repositories related to their codebase. Similarly, finance or HR personnel might regularly work with internal directories containing payroll, compliance, or onboarding documents. These co-access behaviors give rise to group-specific access patterns and interdependencies, forming natural multi-level relationships between users and resources. This hierarchical structure motivates the use of \textbf{graph-based machine learning methods} to model normal behavior and detect deviations indicative of potential security threats.
Traditional UEBA approaches often rely on static baselines or historical data and include machine learning models such as anomaly detectors and LSTM-based deep learning methods~\cite{khaliq2020ueba, names2023ueba}. However, these models frequently require retraining due to shifting behavioral patterns. Other alternatives like particle swarm clustering~\cite{Cui2022} exist, but they also fail to fully leverage multilevel graph structure.
In previous work, we developed a solution combining \textbf{community detection} with \textbf{anomaly detection} on user-entity access patterns, resulting in multiple patents~\cite{US11468029B2, US11468124B2, US11363042B2}. Despite its effectiveness, the approach did not fully exploit the graph hierarchy embedded in user-entity relationships.
To address this, we propose leveraging \textbf{multi-level graph structures} and the theoretical foundation of \textbf{opinion dynamics}~\cite{Ye2021Consensus} to improve anomaly detection. Opinion dynamics, commonly used in influence networks, provides a formal framework to model evolving patterns of trust and deviation among users. This can be adapted to capture access anomalies over time. The goal of this research is to bridge \textbf{opinion dynamics and UEBA} for cloud security by developing an \textbf{adaptive UEBA model} that evolves with changing user behaviors and utilizes underlying relational structures in cloud environments.

\section{Problem Formulation}
\label{sec:problem}

\subsection*{Multi-Level Directory–User Interaction Graph in an Enterprise Setting}
\begin{center}
\includegraphics[width=0.3\textwidth]{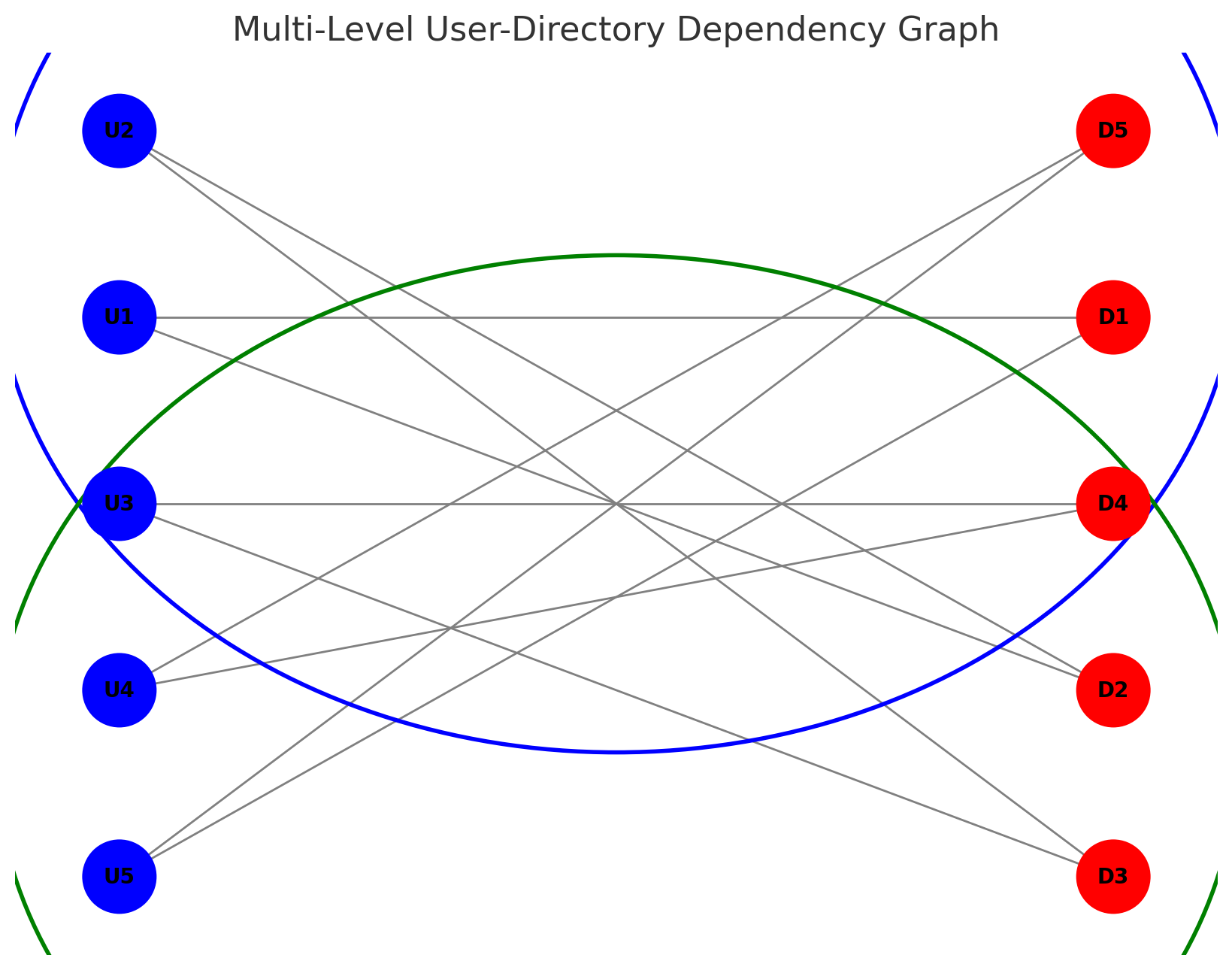}
\end{center}
Consider an enterprise consisting of \( n \geq 2 \) directories \( D_1, D_2, \ldots, D_n \) and \( m \geq 1 \) users \( u_1, u_2, \ldots, u_m \). Users interact with files in these directories through operations such as \texttt{read}, \texttt{write}, \texttt{delete}, or \texttt{update}. These interactions induce access patterns that can be used to infer both directory-to-directory and user-to-user relationships.

\paragraph{Directory-Level Graph \( G[W] \).}

We define a directory similarity graph \( G[W] \), where the adjacency matrix \( W \in \mathbb{R}^{n \times n} \) captures content or behavior-based similarity between directories. The matrix \( W \) is assumed to be \textbf{row-stochastic}:
\[
w_{ii} > 0, \quad \sum_{j=1}^{n} w_{ij} = 1, \quad \forall i \in \{1, \ldots, n\}
\]

This ensures that influence or similarity across directories is normalized, with each directory distributing its influence across all others.


\subsection*{User-User Dependency within a Directory via \( C_i \)}

\begin{figure}[ht]
    \centering
    \begin{tikzpicture}[->, thick, node distance=1.4cm, every node/.style={circle, draw, minimum size=0.8cm}]
        \node[fill=blue!15] (u1) {1};
        \node[fill=blue!15, right of=u1] (u2) {2};
        \node[fill=blue!15, below of=u2] (u3) {3};
        \node[fill=red!15, below of=u1, xshift=-1.5cm] (u4) {4};
        \node[fill=green!15, right of=u3, xshift=1.5cm] (u5) {5};
        \node[fill=green!15, below of=u5] (u6) {6};
        \node[fill=orange!15, below of=u4, xshift=3cm] (u7) {7};

        \path (u1) edge[bend left] (u2)
              (u2) edge[bend left] (u1)
              (u2) edge (u3)
              (u3) edge (u2)
              (u1) edge (u3)
              (u3) edge (u1)
              (u3) edge (u5)
              (u4) edge (u5)
              (u4) edge (u7)
              (u5) edge (u6)
              (u6) edge (u5)
              (u7) edge (u6);
    \end{tikzpicture}
    \caption{User-user logical dependency graph across directories}
    \label{fig:dependency-graph}
\end{figure}
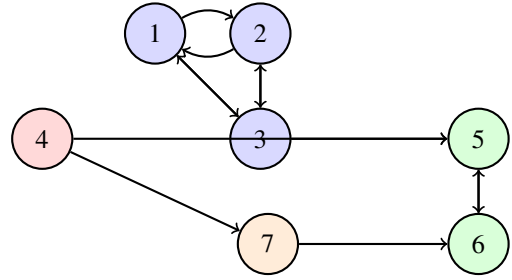

Each directory \( D_i \) is associated with a user-level interaction graph \( G[C_i] \), whose structural form consists of a mixture of open and closed strongly connected components (SCCs), as illustrated in Figure~\ref{fig:dependency-graph}. This structural pattern is assumed to be shared across all directories \( D_1, \ldots, D_n \), though the specific user assignments and edge weights may differ. The interaction graph for each directory is encoded by a matrix \( C_i \in \mathbb{R}^{m \times m} \), where
\begin{align*}
c_{pq,i} \quad &\text{quantifies the logical influence of user } u_q \\
&\text{on user } u_p \text{ in } D_i 
\end{align*}

This matrix \( C_i \) is dynamically updated based on observed access interactions among users within directory \( D_i \). For users not directly interacting within \( D_i \), the corresponding entries in \( C_i \) reflect steady-state values computed from previous interactions, and are periodically updated to reflect long-term behavioral trends.

\paragraph{Strongly Connected Components (SCCs).}

Users naturally cluster into interaction communities (e.g., development teams accessing the same repositories). We represent these as strongly connected components (SCCs) in \( G[C_i] \). These SCCs may be:
\begin{itemize}
    \item \textbf{Closed SCCs:} No incoming edges from other SCCs; internal influence only.
    \item \textbf{Open SCCs:} Receiving influence from users outside the SCC; indicating external behavioral shift or anomaly.
\end{itemize}

\paragraph{Graph Interpretation (See Fig.~\ref{fig:dependency-graph}):}

\begin{itemize}
    \item Users \( \{u_1, u_2, u_3\} \) form a tight, closed SCC with high mutual influence.
    \item Users \( \{u_5, u_6\} \) form a second SCC with balanced bidirectional dependencies.
    \item Users \( u_4 \) and \( u_7 \) show cross-SCC dependencies:
    \begin{itemize}
        \item \( u_4 \to u_5, u_7 \) may suggest access escalation or transition to new roles.
        \item \( u_7 \to u_6 \) indicates cascading influence.
    \end{itemize}
\end{itemize}

\paragraph{Structural Properties (weights) of \( C_i \): Row-Stochasticity and Symmetry.}

The matrix \( C_i \) is assumed to be \textbf{row-stochastic} within each SCC:
\[
\sum_{q=1}^{m} c_{pq,i} = 1, \quad c_{pq,i} \geq 0, \quad \forall p \in \text{SCC}_j(i)
\]

Further, we assume bidirectional influence symmetry within SCCs and no influence between disconnected users:
\[
c_{pq,i} = c_{qp,i}, \& \text{if } u_p, u_q \in \text{same SCC}_j \\
\]

\paragraph{Computing \( C_i \) from Access Data.}

Let \( A_{pq}^i(t) \) denote the observed access-based influence from \( u_q \) to \( u_p \) at time \( t \) for directory \( D_i \). Then each entry of \( C_i(t) \) is computed as:
\[
c_{pq,i}(t) = \frac{A_{pq}^i(t)}{\sum_{k \in \text{CC}_j(i)} A_{pk}^i(t)}
\]

This definition ensures that \( C_i(t) \) is \textbf{normalized per row within each connected component}, and that the influence distribution from each user is bounded and interpretable over time.

\begin{figure}[ht]
    \centering
    \includegraphics[width=0.5\textwidth]{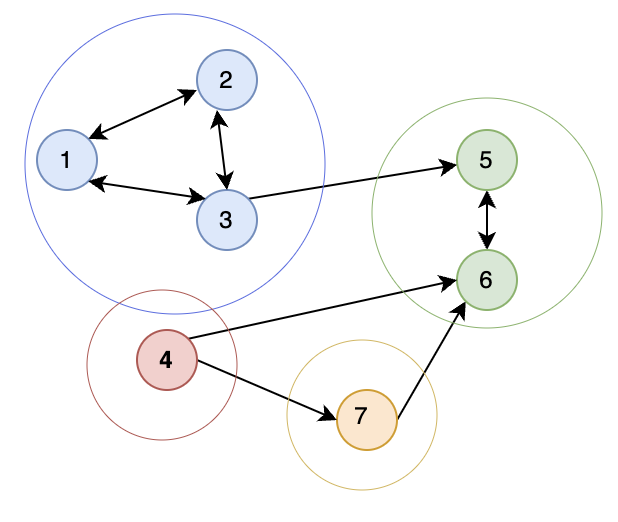}
    \caption{User-user logical dependency graph across directories. Nodes represent users; arrows show influence based on directory-level access behavior. Colored clusters highlight strongly connected components (SCCs).}
    \label{fig:dependency-graph-2}
\end{figure}

\section*{Detecting Anomalies via Multi-Level Interaction Graphs}

We define the **multi-level interaction graph** as \( G(W, C) \), where:
\begin{itemize}
    \item **User Set:** \( U = \{ u_1, u_2, \dots, u_m \} \)
    \item **Directory Set:** \( D = \{ D_1, D_2, \dots, D_n \} \)
    \item **Logical Dependency Graph per Directory \( D_i \):** \( G[C_i] \) represents user-user interactions based on access.
\end{itemize}

If a user \( u_k \) modifies their access behavior at time \( t \), the logical dependency matrix is updated:

\begin{align*}
C_i^t \neq C_i^{t-1} \quad &\text{if user } u_k \text{ accesses a new directory} \\
&\text{or alters behavior.}
\end{align*}

\noindent **Anomalies can be detected by observing:**
\begin{itemize}
    \item Abrupt structural changes in \( C_i^t \) over time.
    \item Deviations from expected user consensus in \( G[C_i] \).
    \item Persistent disagreements in user behavior within strongly connected components.
\end{itemize}

\subsection*{Anomalous Changes in Logical Dependencies}

We consider not only topological changes but also significant internal changes in \( C_i(t) \). An anomaly is flagged if:

\[
\| C_i(t) - C_i(t-1) \|_F > \delta
\]

where \( \| \cdot \|_F \) is the Frobenius norm and \( \delta \) is a sensitivity threshold.

These weight changes impact consensus behavior through:

\[
y(t+1) = \left( W \otimes I_m \right) \cdot \text{vec}(C(t)) \cdot y(t)
\]

\paragraph{Visualization Example:} Suppose \( C_i(t) \) for directory \( D_3 \) at two time steps are:

\[
C_i(t_0) = \begin{bmatrix}
0 & 0.5 & 0.5 \\
0.5 & 0 & 0.5 \\
0.5 & 0.5 & 0
\end{bmatrix}, \quad
C_i(t_1) = \begin{bmatrix}
0 & 0.2 & 0.8 \\
0.4 & 0 & 0.6 \\
0.3 & 0.7 & 0
\end{bmatrix}
\]

The topology remains the same, but internal shifts in influence weights (e.g., from 0.5 to 0.8 or 0.2) suggest behavioral drift. If \( \| C_i(t_1) - C_i(t_0) \|_F > \delta \), we flag \( D_i \) as anomalous.

By continuously monitoring the evolution of \( G(W, C) \), anomalous access behaviors can be detected proactively.

\subsection*{Opinion Dynamics Formulations (Aligned with \cite{Ye2021Consensus})}

We summarize the update rules used in our simulation, each corresponding to the convergence theorems in \cite{Ye2021Consensus}.

\paragraph{Theorem 3: Singleton Topic Dynamics.}
Let \( \mathbf{x}(t) \in \mathbb{R}^n \) be the opinion vector for a singleton topic \( p \). If there are no external dependencies:

\begin{equation}
\mathbf{x}(t+1) = \Gamma_{pp} W \mathbf{x}(t)
\label{eq:singleton_t3}
\end{equation}

where \( \Gamma_{pp} = \mathrm{diag}(c_{pp,1}, \dots, c_{pp,n}) \) and \( W \) is row-stochastic.

\paragraph{Corollary 2.1: Open Singleton Topic.}
If topic \( p \) depends on external topics \( q \in J_p \), each with consensus \( \alpha_q \), then:

\begin{equation}
\mathbf{x}(t+1) = \Gamma_{pp} W \mathbf{x}(t) + \sum_{q \in J_p} \alpha_q \Gamma_{pq}
\label{eq:singleton_open}
\end{equation}

\paragraph{Theorem 3 (Necessity Condition).}
Topic \( p \) reaches consensus \( \alpha_p \) if and only if:

\begin{equation}
\kappa_p (I_n - \Gamma_{pp}) = \sum_{q \in J_p} \alpha_q \Gamma_{pq}, \quad \text{for some } \kappa_p \in [-1,1]
\label{eq:theorem3_kappa}
\end{equation}

Otherwise, consensus is not guaranteed.

\paragraph{Theorem 4: Multi-Topic Open SCC.}
For topics \( \{p_1,\dots,p_r\} \) forming an open SCC:

\begin{equation}
x_i^p(t+1) = c_{pp,i} \sum_j w_{ij} x_j^p(t) + \sum_{q \notin \{p_1,\dots,p_r\}} c_{pq,i} \alpha_q^{(i)}
\label{eq:theorem4-1}
\end{equation}

\paragraph{Theorem 2: Multi-Topic Closed SCC.}
If the SCC is closed and all agents share identical logic over it:

\begin{equation}
x_i^p(t+1) = c_{pp,i} \sum_j w_{ij} x_j^p(t) + \sum_{q \neq p} c_{pq,i} x_i^q(t)
\label{eq:theorem2-1}
\end{equation}

\paragraph{When Does Convergence Fail?}
\begin{itemize}
  \item \textbf{Theorem 2:} Divergence if logic matrices \( C_i \) differ across agents.
  \item \textbf{Theorem 3:} No convergence if Eq.~\eqref{eq:theorem3_kappa} has no solution in \( [-1,1] \).
  \item \textbf{Corollary 2.1:} Divergence if external \( \alpha_q \) values are unresolved or inconsistent.
  \item \textbf{Theorem 4:} Disagreement possible if agents have conflicting logic or differing \( \alpha_q^{(i)} \).
\end{itemize}

\subsection*{Opinion Dynamics Formulations (Aligned with \cite{Ye2021Consensus})}

We present mathematical formulations corresponding to our simulation functions, aligned with Theorems 2–4 from \cite{Ye2021Consensus}.

\paragraph{Singleton Topic Dynamics (Theorem 3).}
Let \( \mathbf{x}(t) \in \mathbb{R}^n \) represent the opinions of \( n \) agents on a singleton topic \( p \in J \) at time \( t \). The evolution is given by:
\begin{equation}
\mathbf{x}(t+1) = \Gamma_{pp} \cdot W \cdot \mathbf{x}(t)
\label{eq:singleton_theorem3}
\end{equation}
where:
\begin{itemize}
    \item \( W \in \mathbb{R}^{n \times n} \) is a row-stochastic influence matrix,
    \item \( \Gamma_{pp} = \mathrm{diag}(c_{pp,1}, \ldots, c_{pp,n}) \) is the diagonal matrix of self-dependencies for topic \( p \).
\end{itemize}
Under strong connectivity and aperiodicity of \( W \), Theorem 3 ensures convergence to a steady state \( \mathbf{x}^* \in [-1,1]^n \).

\paragraph{Open Singleton Topic with External Dependencies (Corollary 2.1).}
When topic \( p \) depends on a set of external topics \( J_p \subset J \), each with known consensus \( \alpha_q \), the update becomes:
\begin{equation}
\mathbf{x}(t+1) = \Gamma_{pp} \cdot W \cdot \mathbf{x}(t) + \sum_{q \in J_p} \alpha_q \cdot \Gamma_{pq}
\label{eq:singleton_corollary21}
\end{equation}
where \( \Gamma_{pq} = \mathrm{diag}(c_{pq,1}, \ldots, c_{pq,n}) \) represents logical influence of topic \( q \) on topic \( p \).

\paragraph{Multi-Topic Open SCC Dynamics (Theorem 4).}
Let \( \mathbf{X}(t) \in \mathbb{R}^{n \times r} \) be the opinion matrix of \( n \) agents on \( r \) interdependent topics \( \{p_1, \ldots, p_r\} \subset J \), forming an open SCC. The evolution for agent \( i \) and topic \( p \in \{p_1,\ldots,p_r\} \) is:
\begin{equation}
x_i^p(t+1) = c_{pp,i} \sum_{j=1}^n w_{ij} x_j^p(t) + \sum_{q \notin \{p_1,\dots,p_r\}} c_{pq,i} \, \alpha_q^{(i)}
\label{eq:theorem4-2}
\end{equation}
with:
\begin{itemize}
    \item \( c_{pp,i} \) and \( c_{pq,i} \) from agent-specific logic matrix \( C_i \),
    \item \( \alpha_q^{(i)} \) representing scalar or vector consensus values for external topics,
    \item \( x_i^p(t) \) is the opinion of agent \( i \) on topic \( p \) at time \( t \).
\end{itemize}
This corresponds to Equation (14) in Theorem 4 of \cite{Ye2021Consensus} and supports modeling of joint logical influence and external dependency resolution.

\paragraph{Multi-Topic Closed SCC Dynamics (Theorem 2).}
Let \( \mathbf{X}(t) \in \mathbb{R}^{n \times r} \) be the matrix of opinions over a closed, strongly connected set of topics \( \{p_1, \ldots, p_r\} \). If all agents share the same logic submatrix (i.e., \( C_i|_{\text{SCC}} = C_j|_{\text{SCC}} \) for all \( i, j \)), the update rule is:
\begin{equation}
x_i^p(t+1) = c_{pp,i} \sum_{j=1}^n w_{ij} x_j^p(t) + \sum_{q \in \{p_1,\ldots,p_r\} \setminus \{p\}} c_{pq,i} \cdot x_i^q(t)
\label{eq:theorem2-2}
\end{equation}
This models internal logical coupling among the topics within the SCC, and under the assumptions of Theorem 2 from \cite{Ye2021Consensus}, the opinions \( \mathbf{X}(t) \) converge to topic-wise consensus.

\paragraph{Conditions Leading to Divergence or Disagreement (from \cite{Ye2021Consensus}):}

\begin{itemize}
    \item \textbf{Theorem 2 (Multi-topic Closed SCC)} requires all agents to share the same logic submatrix over the topics in the SCC.\\
    \textit{Violation:} If \( C_i|_{\text{SCC}} \neq C_j|_{\text{SCC}} \) for some \( i \neq j \), then consensus is not guaranteed; topic-wise disagreement may persist.

    \item \textbf{Theorem 3 (Singleton Topic)} assumes the topic has no logical dependency on other topics and that the self-dependency values \( c_{pp,i} \in (0,1] \) are strictly positive.\\
    \textit{Violation:} If \( c_{pp,i} = 0 \) for any agent \( i \), or if \( W \) is not strongly connected and aperiodic, convergence may fail.

    \item \textbf{Corollary 2.1 (Open Singleton Topic)} requires external topics \( q \in J_p \) to have already converged to consensus values \( \alpha_q \).\\
    \textit{Violation:} If any dependent topic \( q \) has not reached consensus or exhibits disagreement, topic \( p \) may not converge.

    \item \textbf{Theorem 4 (Multi-topic Open SCC)} allows different agents to have different logic matrices. However, convergence requires:
    \begin{enumerate}
        \item The SCC is closed with respect to external dependencies once \( \alpha_q \) values are injected,
        \item The joint system (logic and influence) has bounded self-dependencies and stable external inputs.
    \end{enumerate}
    \textit{Violation:} If external consensus values \( \alpha_q^{(i)} \) are inconsistent across agents or if agents have conflicting logic structures (e.g., competing interdependencies), the SCC may exhibit disagreement or oscillatory behavior.
\end{itemize}








\section{Main Results}

\subsection{SCC Decomposition and Theorem Assignment}

We adopt the decomposition and evaluation framework from \cite{ye2018consensus}, where the topic index set \( \mathcal{J} \) is partitioned into disjoint blocks \( \mathcal{J}_j \), each corresponding to a strongly connected component (SCC) in the logical dependency graph. This is formalized as:
\[
\mathcal{J}_j \triangleq \left\{ \sum_{i=1}^{j} s_{i-1} + 1, \sum_{i=1}^{j} s_{i-1} + 2, \dots, \sum_{i=1}^{j} s_{i-1} + s_j \right\}
\]
where \( s_j \) denotes the number of topics in SCC \( \mathcal{J}_j \).

\subsection{Decomposition and Evaluation Pipeline}

To analyze opinion dynamics in a modular and theoretically justified manner, we follow these steps:

\begin{enumerate}
    \item \textbf{SCC Extraction:} For each logic matrix \( C_i \), build a directed graph and extract all SCCs.
    \item \textbf{Open/Closed Classification:} Mark \( \mathcal{J}_j \) as \textit{open} if any topic depends on external topics; otherwise, mark it \textit{closed}.
    \item \textbf{Local Dependencies:} For each topic \( p \in \mathcal{J}_j \), define
    \[
    \hat{J}_p = \{ q \neq p : C[p, q] \neq 0 \}
    \]
    \item \textbf{External Dependencies:} Set \( \tilde{J}_j = \bigcup_{p \in \mathcal{J}_j} \hat{J}_p \setminus \mathcal{J}_j \)
    \item \textbf{Dependency Conditions:} Construct a DAG over SCCs; evaluate each block only after its DAG predecessors.
    \item \textbf{Theorem Assignment:} Based on the dependency structure and numerical values in \( C_i \), assign Theorem 2, 3, Corollary 2.1, or 4.
\end{enumerate}

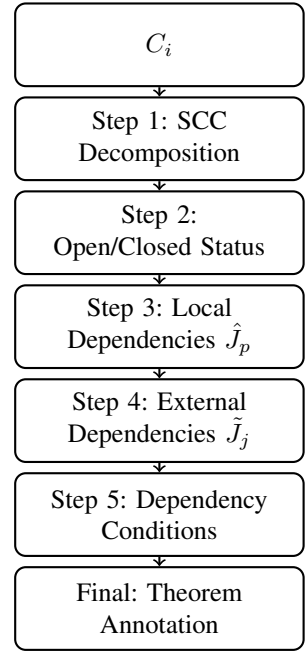
\begin{figure}[ht]
    \centering
    \begin{tikzpicture}[node distance=1.25cm, every node/.style={draw, rectangle, rounded corners, minimum height=1.1cm, align=center, text width=3.6cm}, every path/.style={->, thick}]
        \node (Ci) {\textbf{$C_i$}};
        \node (scc) [below of=Ci] {Step 1: SCC Decomposition};
        \node (status) [below of=scc] {Step 2: Open/Closed Status};
        \node (local) [below of=status] {Step 3: Local Dependencies $\hat{J}_p$};
        \node (external) [below of=local] {Step 4: External Dependencies $\tilde{J}_j$};
        \node (condition) [below of=external] {Step 5: Dependency Conditions};
        \node (theorem) [below of=condition] {Final: Theorem Annotation};

        \draw[->] (Ci) -- (scc);
        \draw[->] (scc) -- (status);
        \draw[->] (status) -- (local);
        \draw[->] (local) -- (external);
        \draw[->] (external) -- (condition);
        \draw[->] (condition) -- (theorem);
    \end{tikzpicture}
    \caption{Pipeline for SCC decomposition and theorem annotation.}
    \label{fig:scc-pipeline}
\end{figure}

\subsection{Example: Theorem Mapping for Paper Simulation}

\begin{figure}[ht]
    \centering
    \includegraphics[width=0.5\textwidth]{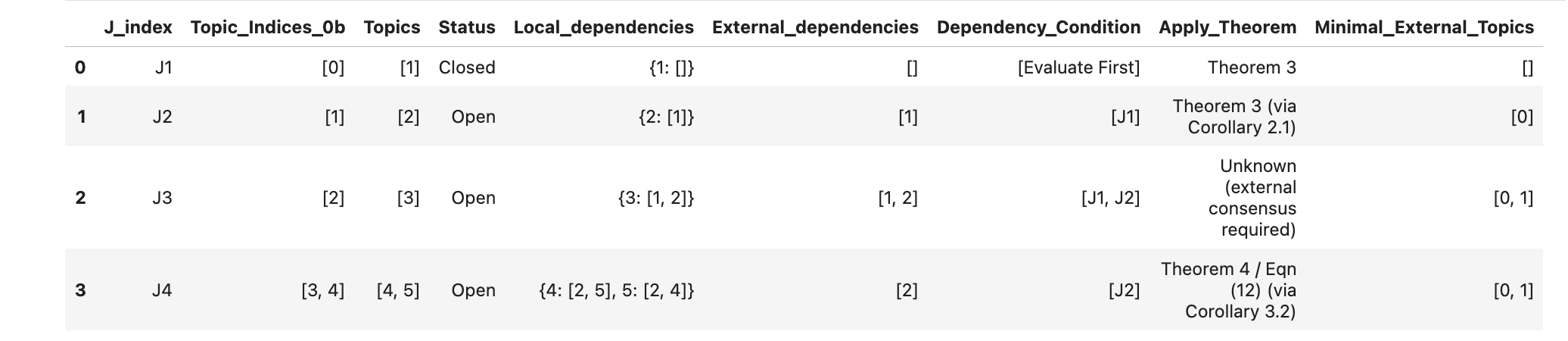}
    \caption{Annotated SCC blocks with assigned theorems, based on logical structure and dependency ordering in \cite{ye2018consensus}.}
    \label{fig:theorem-annotation-img}
\end{figure}

The table  in \autoref{fig:theorem-annotation-img},
 replicates the SCC analysis for the simulation in \cite{ye2018consensus} by authost themselves, Reference 29, aligning each topic block with its evaluation order and applicable theorem.

\subsection{Main Algorithm: Bayesian Detection of Anomalies via Opinion Variance Shifts}
\label{sec:bayesian-anomaly}
\begin{algorithm}[tb]
\caption{Bayesian Anomaly Score Estimation via Scaled Variance with Online Prior Update}
\label{alg:anomaly}
\begin{algorithmic}[1]
\State \textbf{Input:} $x_{\text{prev}} \in \mathbb{R}^{n \times d}$, $x_{\text{now}} \in \mathbb{R}^{n \times d}$, prior $\pi_0$, scale factor $s$, exponent $\alpha$
\State \textbf{Output:} Anomaly score $\pi \in [0, 1]$

\If{$x_{\text{prev}}$ is 1D}
    \State $x_{\text{prev}} \gets \text{reshape}(x_{\text{prev}})$
\EndIf
\If{$x_{\text{now}}$ is 1D}
    \State $x_{\text{now}} \gets \text{reshape}(x_{\text{now}})$
\EndIf

\State $\text{Var}_{\text{cur}} \gets \operatorname{Var}(s \cdot x_{\text{now}})$ \Comment{Current variance (scaled)}
\State $\text{Var}_{\text{prev}} \gets \operatorname{Var}(s \cdot x_{\text{prev}})$
\State $v_{\text{cur}} \gets \frac{1}{d} \sum_j \text{Var}_{\text{cur}}[j]$
\State $v_{\text{prev}} \gets \frac{1}{d} \sum_j \text{Var}_{\text{prev}}[j]$
\State $\Delta v \gets \max(v_{\text{cur}} - v_{\text{prev}}, 0)$
\State $L \gets 1 - \exp(-\alpha \cdot \Delta v)$ \Comment{Exponential likelihood mapping}

\State $\pi_0 \gets \pi_{t-1}$ \Comment{Posterior from previous time}
\State $\pi \gets \dfrac{L \cdot \pi_0}{L \cdot \pi_0 + (1 - L)(1 - \pi_0)}$ \Comment{Bayesian update}

\State \Return $\pi$
\end{algorithmic}
\end{algorithm}

The proposed anomaly detection algorithm builds upon the structure of opinion dynamics in multi-agent logical systems. It assumes that under stable conditions, opinions converge (or exhibit bounded disagreement) as characterized by Theorems 2–4 in \cite{Ye2021Consensus}. Deviations from these behaviors are flagged as anomalies, especially when accompanied by a rise in opinion variance.

\paragraph{Variance as a Signature of Disruption.}
We use the per-topic opinion variance across agents as a signal of stability or disruption. Under normal operation—especially in closed SCCs (Theorem 2 or 3)—variance remains low and stable. A sudden spike in:
\[
\Delta v = \max(v_{\text{cur}} - v_{\text{prev}}, 0)
\]
indicates that the system is deviating from expected behavior, often due to logic inconsistency, agent drift, or injection of anomalous entities.

\paragraph{Change in Self-Dependencies \( c_{pp,i} \).}
A key cause of increased variance is a change in the diagonal logic weights \( c_{pp,i} \), which govern how much agent \( i \) relies on external opinions. In our setting, these are modeled as:
\[
c_{pp,i} \propto a_{ii}
\]
where \( a_{ii} \) represents the normalized self-access intensity of user \( i \) within a directory (e.g., read/write frequency). An increase in \( a_{ii} \) makes agent \( i \) more self-reliant, thereby dampening consensus behavior and causing divergence from the group. This shift directly contributes to higher opinion variance and will be clearly reflected in simulation outputs.

\paragraph{Exponential Likelihood Mapping and Bayesian Update.}
To detect anomalies, we use a scaled exponential likelihood:
\[
L = 1 - \exp(-\alpha \cdot \Delta v)
\]
and apply a Bayesian update:
\[
\pi_t = \frac{L \cdot \pi_{t-1}}{L \cdot \pi_{t-1} + (1 - L)(1 - \pi_{t-1})}
\]
where \( \pi_{t-1} \) is the prior anomaly probability and \( \pi_t \) is the posterior belief. This enables smooth, probabilistic anomaly tracking over time.

\paragraph{Connection to Simulations.}
In our simulation framework, changes in \( c_{pp,i} \) values due to increased \( a_{ii} \), or structural disagreements (e.g., new logical edges or SCC dependencies), both manifest as observable increases in opinion variance. The anomaly score \( \pi_t \) thus encapsulates both local logic shifts and global structural instabilities.

This allows our model to:
\begin{itemize}
    \item React to local agent behavior change (via \( c_{pp,i} \)),
    \item Detect structural disagreement (via Theorem 3/4 outputs),
    \item Update anomaly likelihoods recursively and online.
\end{itemize}

\section{Simulations and Experiments}

\subsection{Simulation 1: Reproducing the Original Paper's Setup}

To validate our implementation, we replicated the simulation described by the authors in \cite{ye2018consensus}. The simulation involves a network of \( n = 6 \) agents interacting over 5 logically coupled topics using the influence matrix \( W \) and topic logic matrices \( \hat{C} \), \( \bar{C} \), and \( \tilde{C} \). These matrices are provided directly as follows:

\paragraph{Influence Matrix \( W \):}
\[
W = \begin{bmatrix}
0.2 & 0 & 0 & 0 & 0.8 & 0 \\
0.5 & 0.3 & 0 & 0 & 0 & 0.2 \\
0 & 0.3 & 0.1 & 0 & 0 & 0.6 \\
0 & 0 & 0.85 & 0.15 & 0 & 0 \\
0 & 0 & 0 & 0.2 & 0.8 & 0 \\
0 & 0 & 0 & 0 & 0.5 & 0.5
\end{bmatrix}
\]

\paragraph{Logic Matrices (Different Agent Beliefs):}

\[
\hat{C} =
\begin{bmatrix}
1 & 0 & 0 & 0 & 0 \\
-0.5 & 0.5 & 0 & 0 & 0 \\
-0.3 & -0.6 & 0.1 & 0 & 0 \\
0 & -0.3 & 0 & 0.2 & -0.5 \\
0 & -0.5 & 0 & -0.2 & 0.3
\end{bmatrix}
\]
\[
\bar{C} =
\begin{bmatrix}
1 & 0 & 0 & 0 & 0 \\
-0.8 & 0.2 & 0 & 0 & 0 \\
-0.3 & -0.1 & 0.6 & 0 & 0 \\
0 & -0.3 & 0 & 0.2 & -0.5 \\
0 & -0.5 & 0 & -0.2 & 0.3
\end{bmatrix}
\]

\paragraph{Perturbed Matrix \( \tilde{C} \):}
This variant uses altered logic entries to create divergence conditions:
\[
\tilde{C} =
\begin{bmatrix}
1 & 0 & 0 & 0 & 0 \\
0.5 & 0.5 & 0 & 0 & 0 \\
-0.3 & -0.1 & 0.6 & 0 & 0 \\
0 & -0.3 & 0 & 0.2 & -0.5 \\
0 & -0.5 & 0 & -0.2 & 0.3
\end{bmatrix}
\]

\subsubsection*{Opinion Dynamics Results for Different \( C \) Matrices}

We validated the correctness of our implementation by comparing simulated outcomes to the expected results as stated in the paper. The table below summarizes which topics converge correctly under each logic matrix:

\begin{figure}[ht]
    \centering
    \includegraphics[width=0.55\textwidth]{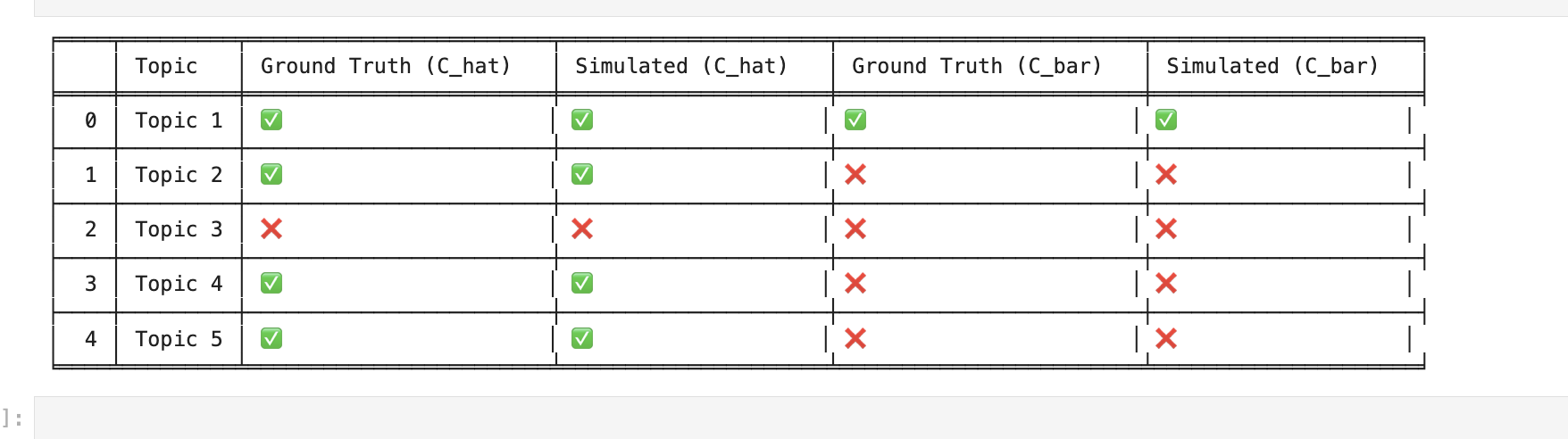}
    \caption{Ground truth validation for Topics 1–5 under logic matrices \( \hat{C} \) and \( \bar{C} \). Topic 3 fails to converge as expected under \( \bar{C} \).}
    \label{fig:gt-table}
\end{figure}

\begin{figure}[ht]
    \centering
    \begin{subfigure}[b]{0.48\textwidth}
        \centering
        \includegraphics[width=\textwidth]{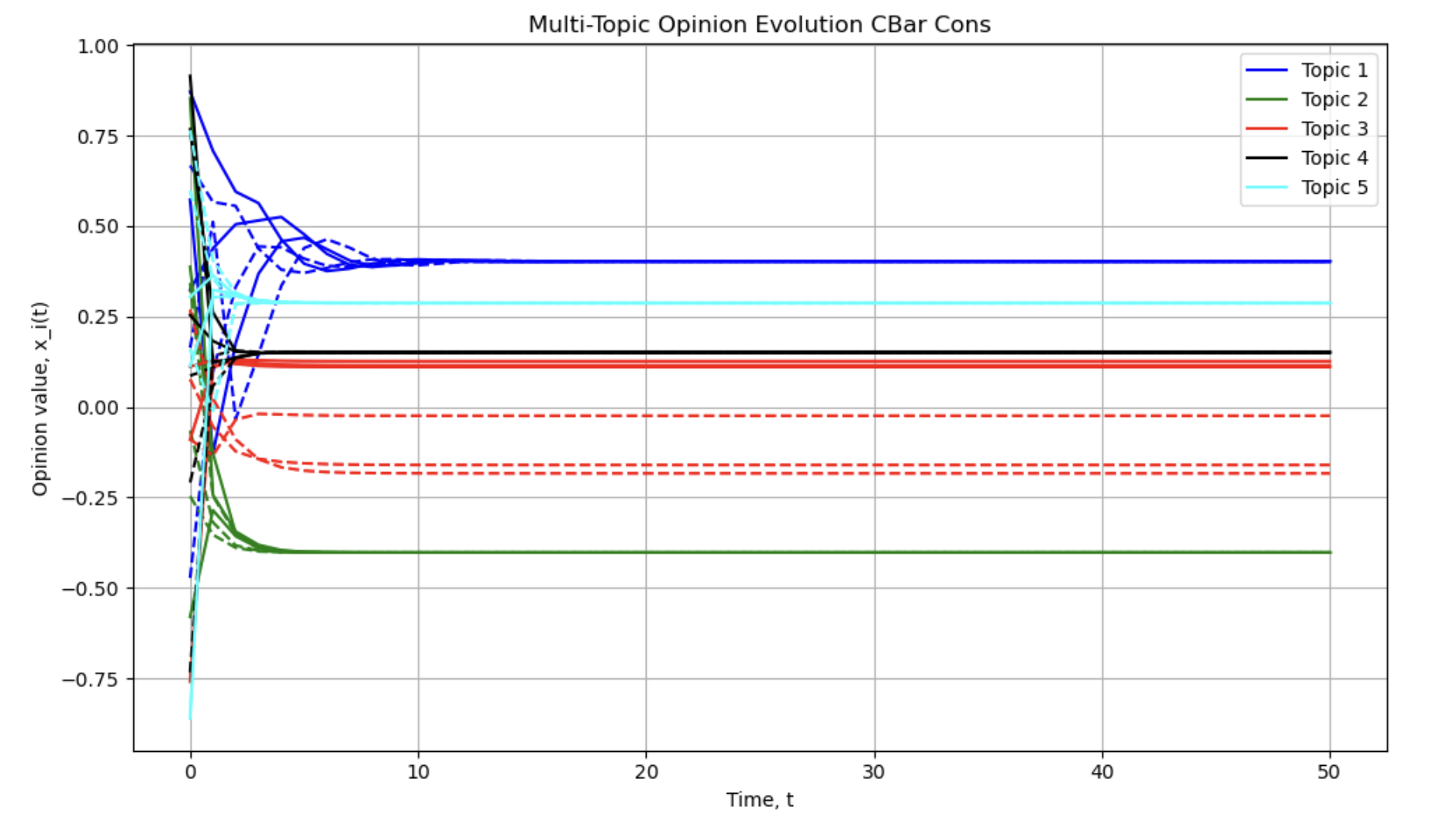}
        \caption{Opinion evolution over 50 steps with logic matrix \( \bar{C} \). Most topics stabilize, but some agents diverge on Topic 3.}
        \label{fig:sim-cbar}
    \end{subfigure}
    \hfill
    \begin{subfigure}[b]{0.48\textwidth}
        \centering
        \includegraphics[width=\textwidth]{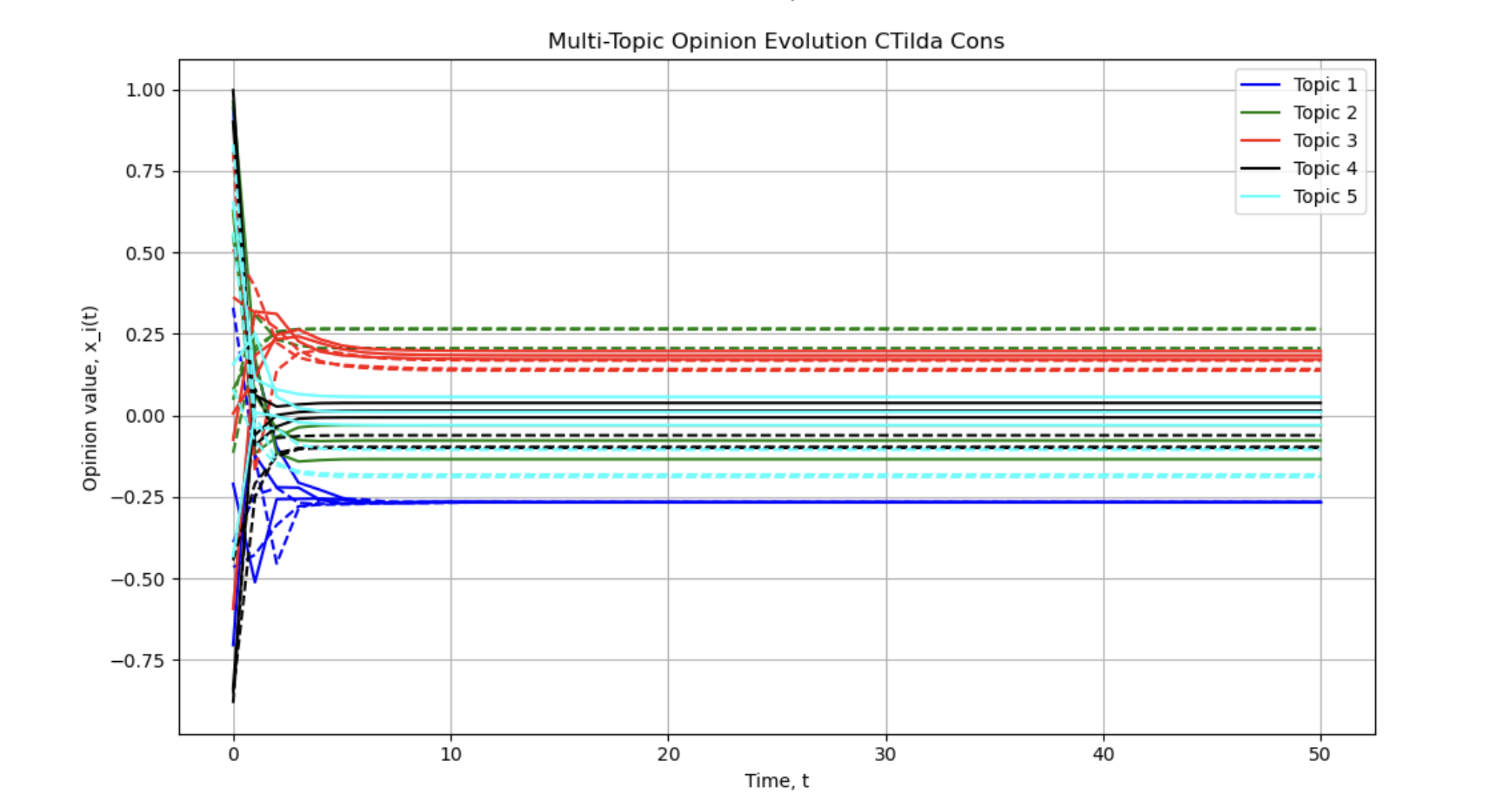}
        \caption{Opinion evolution with perturbed matrix \( \tilde{C} \). Increased divergence due to inconsistent logic between agents.}
        \label{fig:sim-ctilde}
    \end{subfigure}
    \caption{Comparison of opinion evolution under two logic configurations.}
    \label{fig:sim-comparison}
\end{figure}

These graphs in \ref{fig:sim-cbar} \ref{fig:sim-ctilde} match  simulation present in the paper \cite{ye2018consensus}

For implementation details, refer to my open-source code at \url{https://github.com/SankarshU/Graph-Machine-Learning/tree/5aec0a917ef346bef1a698f0dab3d575cb118c4d/Consenus}.




\subsection{Bayesian Anomaly Detection: Design, Dynamics, and Inference}

We model a cloud access behavior system where each directory is represented as a topic and users as agents. Agents evolve their opinions over time using logic dependency matrices \( C_i \) and a shared influence matrix \( W \), governed by the consensus dynamics model of Ye et al.~\cite{Ye2021Consensus}. Anomalous behavior is simulated by modifying the structure of inter-topic logical dependencies at time \( T=5 \).

\paragraph{Influence Matrix \( W \):}
\[
\resizebox{\linewidth}{!}{
$
W =
\begin{bmatrix}
0.281 & 0.334 & 0.334 & 0 & 0 & 0.050 & 0 \\
0.333 & 0.333 & 0.333 & 0 & 0 & 0 & 0 \\
0.317 & 0.317 & 0.317 & 0 & 0 & 0.048 & 0 \\
0.050 & 0 & 0 & 0.450 & 0.500 & 0 & 0 \\
0.050 & 0 & 0 & 0.450 & 0.500 & 0 & 0 \\
0.100 & 0.050 & 0 & 0.100 & 0.050 & 0.650 & 0.050 \\
0 & 0 & 0 & 0 & 0 & 0.050 & 0.950
\end{bmatrix}
$
}
\]
\textit{\small D1 → small influence from D6, D2 unchanged, D3 sees D6, D4/D5 influenced by D1, D6 touches all, D7 receives only from D6.}

\vspace{0.5em}
\paragraph{Logic Matrix \( \hat{C} \) (Baseline, Row-normalized):}
\[
\hat{C} =
\begin{bmatrix}
0.143 & 0.429 & 0.429 & 0 & 0 & 0 & 0 \\
0.429 & 0.143 & 0.429 & 0 & 0 & 0 & 0 \\
0.429 & 0.429 & 0.143 & 0 & 0 & 0 & 0 \\
0 & 0 & 0 & 0.667 & 0.333 & 0 & 0 \\
0 & 0 & 0 & 0.333 & 0.667 & 0 & 0 \\
0 & 0 & 0 & 0 & 0 & 1 & 0 \\
0 & 0 & 0 & 0 & 0 & 0 & 1 \\
\end{bmatrix}
\]

We depict a simulated cloud system where users are modeled as agents and directories as topics. Each directory has an associated logic matrix \( C_i \), representing topic-topic dependencies based on user activity. The influence matrix \( W \) governs inter-directory similarity.

\paragraph{Steady-State Structure at \( T = 0 \):}

At initialization, each directory (D1–D7) uses the same logic matrix \( C_{\text{hat}} \), forming closed strongly connected components (SCCs):

\begin{itemize}
    \item D1–D3: Mutual influence among Users 1, 2, 3.
    \item D4–D5: Bi-directional mutual influence between Users 4 and 5.
    \item D6, D7: Isolated users with self-loop only.
\end{itemize}

\begin{figure}[ht]
    \centering
    \includegraphics[width=0.5\textwidth]{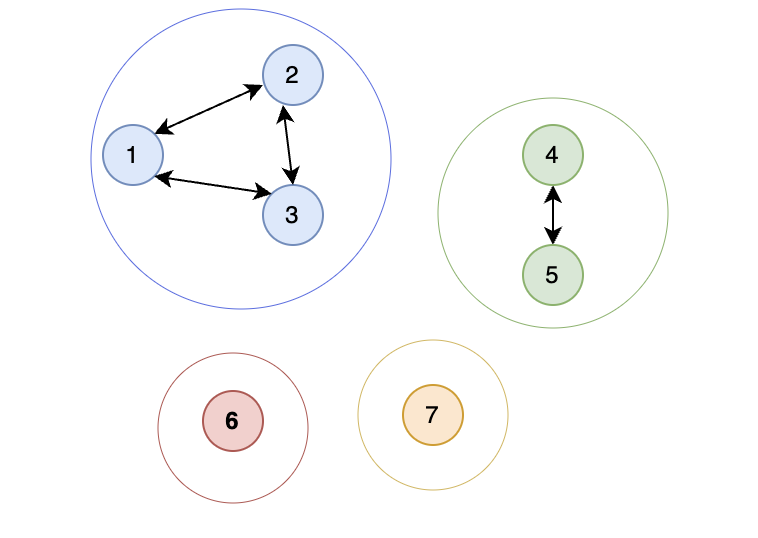} 
    \caption{Initial SCC decomposition at \( T=0 \): no cross-component influences. All directories use logic matrix \( C_{\text{hat}} \).}
    \label{fig:t0-structure}
\end{figure}

\paragraph{Anomaly Injection at \( T = 5 \):}

At \( T = 5 \), we simulate an abnormal access event: Users 2 and 3 in D1–D3 begin influencing D4 and D5 (Users 4 and 5) with tunable weight \( \text{wt} \). The logic matrices of D4 and D5 are modified and row-normalized to reflect this structural anomaly, producing a new matrix \( C_{\text{bar}} \).

\begin{figure}[ht]
    \centering
    \includegraphics[width=0.55\textwidth]{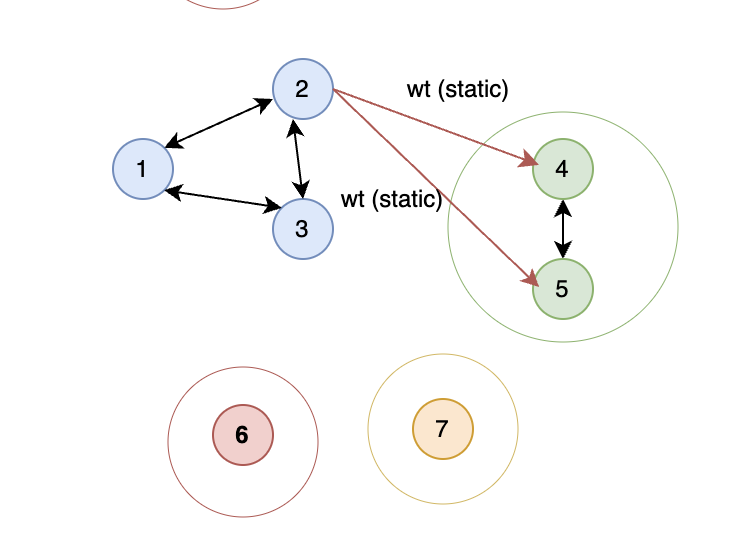} 
    \caption{Abnormal access injected at \( T=5 \): static cross-component influence from User 2 to Users 4 and 5. Affected directories switch to logic matrix \( C_{\text{bar}} \).}
    \label{fig:t5-static}
\end{figure}

\paragraph{Weighted Anomaly Variation:}

We vary the influence weight \(\text{wt} \in \{1, 2, 5, 10, \ldots\}\) to study its effect on divergence. Each setting results in a different logic matrix \( C_{\text{bar}}^{(\text{wt})} \). For large weights, the influence of abnormal users dominates and alters D4/D5 behavior significantly.

\begin{figure}[ht]
    \centering
    \includegraphics[width=0.55\textwidth]{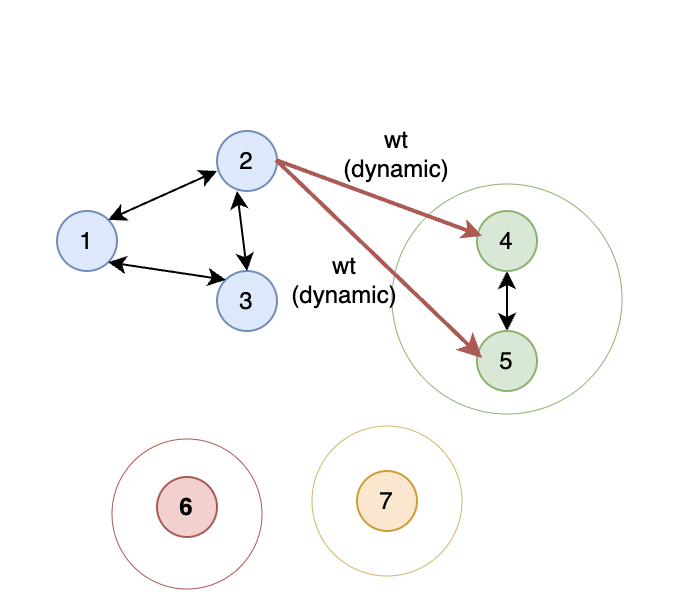} 
    \caption{Illustration of dynamic weighted influence: cross-component edges from User 2 to D4/D5 are varied. Logic matrix becomes \( C_{\text{bar}}^{(\text{wt})} \).}
    \label{fig:t5-dynamic}
\end{figure}

\paragraph{Key Notes:}
\begin{itemize}
    \item The rest of the directories (D1–D3, D6, D7) retain their original logic structure \( C_{\text{hat}} \).
    \item The simulation pipeline checks the effect of anomaly injection on opinion evolution using Theorem 3 and scaled variance drift.
\end{itemize}

\vspace{0.5em}
\paragraph{Perturbed Matrix \( \bar{C} \) at \( w_t = 2 \):}
\[
\bar{C}^{(w=2)} =
\begin{bmatrix}
\cdots \\
\cdots \\
\mathbf{0} & \mathbf{0.571} & \mathbf{0} & \mathbf{0.143} & \mathbf{0.286} & \cdots \\
\mathbf{0} & \mathbf{0.571} & \mathbf{0} & \mathbf{0.286} & \mathbf{0.143} & \cdots \\
\cdots \\
\end{bmatrix}
\]

\vspace{0.5em}
\paragraph{Perturbed Matrix \( \bar{C} \) at \( w_t = 50 \):}
\[
\bar{C}^{(w=50)} =
\begin{bmatrix}
\cdots \\
\cdots \\
\mathbf{0} & \mathbf{0.971} & \mathbf{0} & \mathbf{0.010} & \mathbf{0.019} & \cdots \\
\mathbf{0} & \mathbf{0.971} & \mathbf{0} & \mathbf{0.019} & \mathbf{0.010} & \cdots \\
\cdots \\
\end{bmatrix}
\]

\textit{\small Bold rows highlight the abnormal influence injected from User 1 to D4, D5. As \( w_t \) increases, the normalized logic weights heavily shift toward user 1.}

\paragraph{Setup at \( T=0 \):}
\begin{itemize}
    \item All directories (topics) begin with stable logical dependency matrices \( C_i \), where subsets of users form closed strongly connected components (SCCs).
    \item Steady state consensus is reached per directory by simulating opinion dynamics for 5000 steps.
    \item No anomalous cross-directory access is introduced yet. The system converges with all SCCs and topics stabilized.
\end{itemize}

\paragraph{Perturbation at \( T=5 \):}
\begin{itemize}
    \item We inject an abnormal influence from user/topic 1 into directories D3 and D4 by modifying specific entries of the \( C_i \) matrix.
    \item The updated matrices, row-normalized, reflect strong cross-SCC influence, altering only topics 3 and 4.
    \item Remaining directories retain their structure:
    \[
    \texttt{C\_list\_bar} = [C_{\text{bar}}^{(1)}] * 3 + [C_{\text{bar}}^{(2)}] * 2 + [C_{\text{bar}}^{(1)}] * 2
    \]
\end{itemize}

\begin{figure}[ht]
    \centering
    \includegraphics[width=0.99\linewidth]{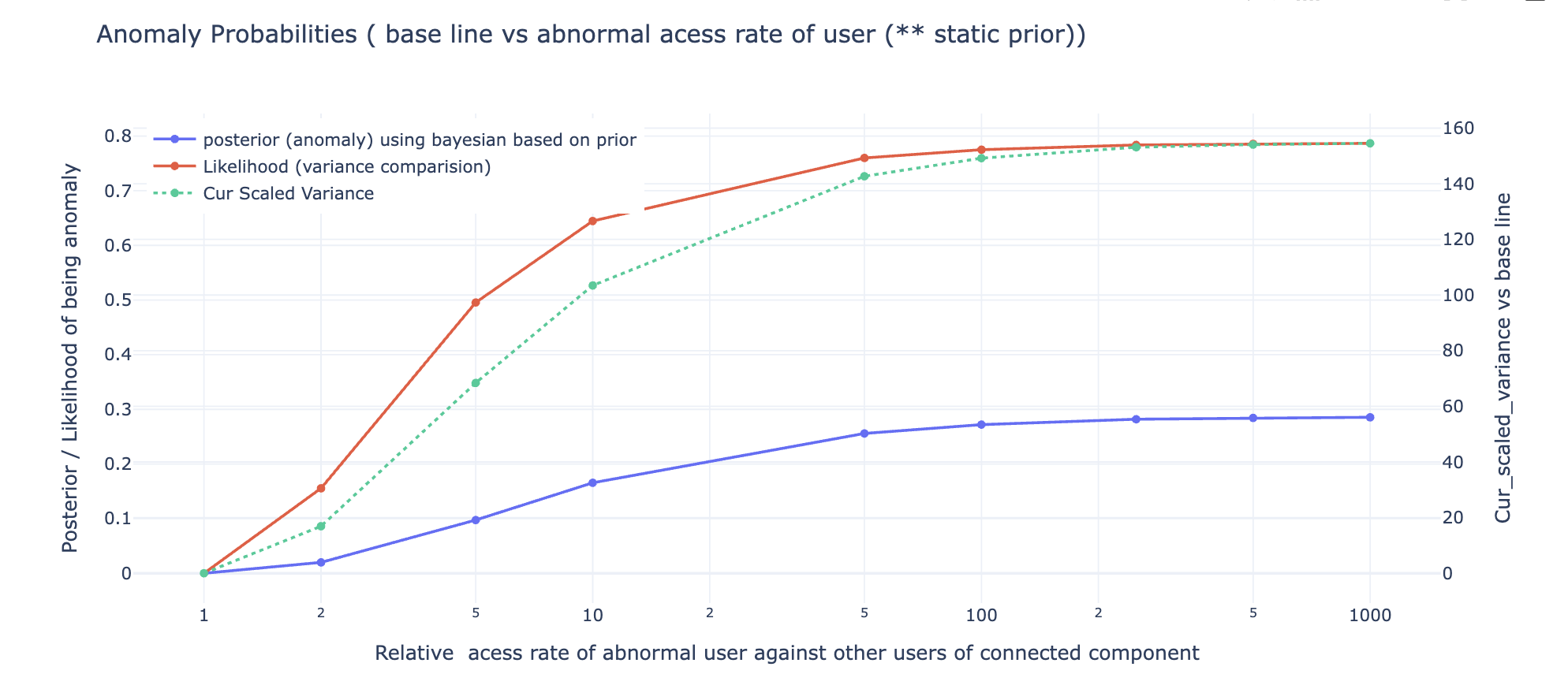} \\\vspace{0.5em}
    \includegraphics[width=0.99\linewidth]{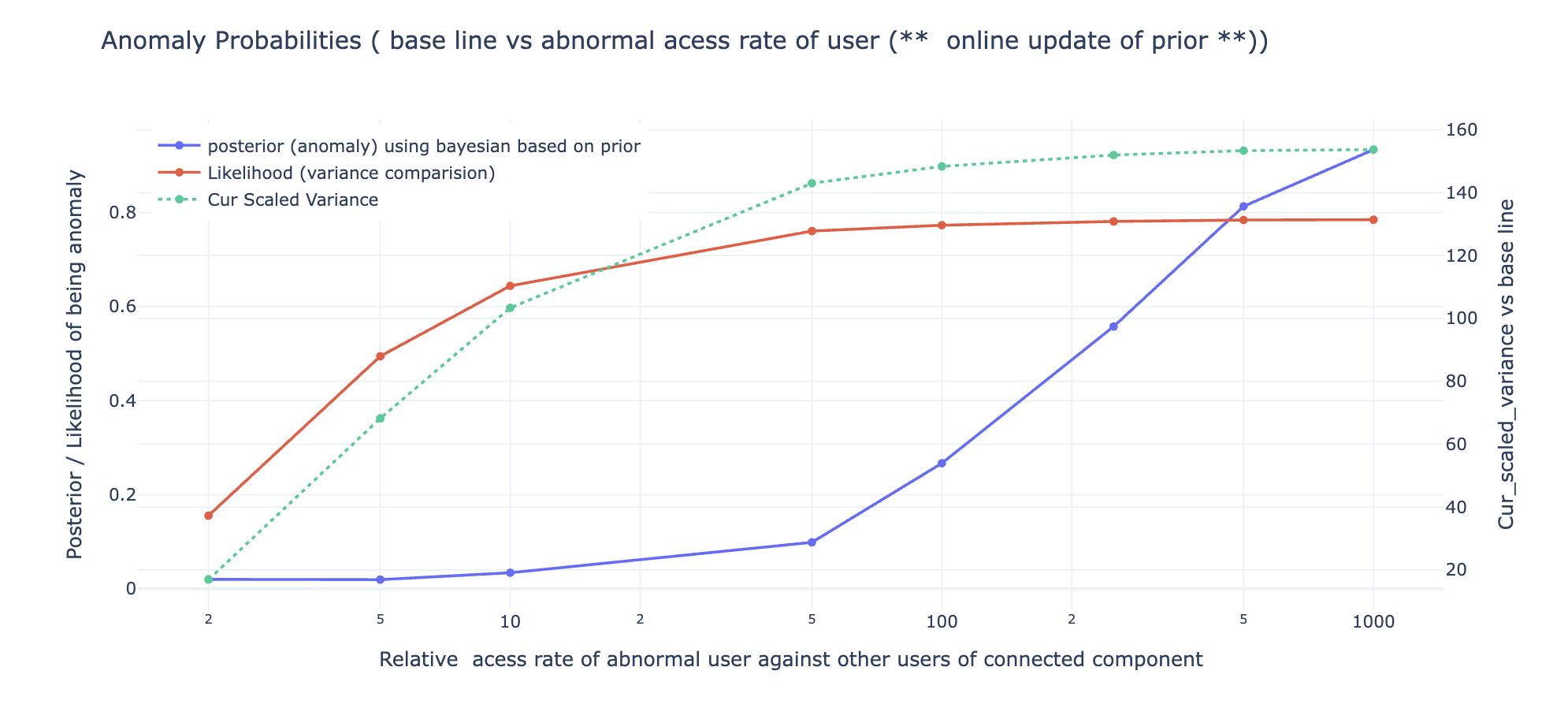}
    \caption{
        \textbf{Bayesian Anomaly Detection under Static vs Online Prior Updates.}
        Top: Static prior (\( \pi_0 = 0.1 \)) kept constant. Bottom: Online prior update via posterior at each step.
        Increasing anomaly weights cause rising variance and anomaly probability.
        Online inference detects shifts earlier and reacts more sharply.
    }
    \label{fig:bayesian-prior-comparison}
\end{figure}

\paragraph{Detection Pipeline:}
\begin{itemize}
    \item For each influence weight \( w_t \in \{1, 2, 5, 10, \ldots, 1000\} \), we compare the current opinion matrix \( x_{\text{now}} \) with the baseline \( x_{\text{prev}} \).
    \item A likelihood score is derived from the shift in scaled topic variance.
    \item An anomaly score is computed using a Bayesian update. Two scenarios are evaluated:
    \begin{enumerate}
        \item \textbf{Static prior:} fixed anomaly prior \( \pi_0 = 0.1 \)
        \item \textbf{Online prior:} \( \pi_t \leftarrow \text{posterior}_{t-1} \)
    \end{enumerate}
\end{itemize}

This setup captures both structural deviation (via modified logic graphs) and behavioral response (via probabilistic scoring), enabling a fine-grained analysis of abnormal access propagation in interconnected opinion networks.The anomaly score for different wights can be observed in  Figure~\ref{fig:bayesian-prior-comparison}, ...

\subsection{Observations}
\begin{itemize}
    \item As the influence weight \( w_t \) increases, both scaled variance and likelihood of anomaly increase.
    \item Under static prior, posterior saturates early—less sensitive to cumulative divergence.
    \item Under online updating, posterior compounds over time, enabling more responsive detection at lower weights.
\end{itemize}

\section{CHALLENGES, CONCLUSION, AND NEXT STEPS}

\subsection{Challenges of The Project.}
Our implementation highlights several critical challenges:
First and foremost, faithfully replicating the opinion dynamics framework from~\cite{ye2018consensus} proved non-trivial. The model requires recursive decomposition of logical matrices into strongly connected components (SCCs), dynamic dependency resolution, and conditional theorem application depending on the structure and external influence. Ensuring correctness in this modular yet tightly-coupled architecture is highly susceptible to subtle bugs.

\subsection{Open issues.}
\begin{itemize}
    \item \textbf{Scalability:} As the number of users and directories grows into the thousands, matrix-based simulations, consensus propagation, and dependency checks become compute-intensive. Optimization via lazy updates, SCC caching, or distributed processing is essential.

    \item \textbf{False Positives:} In real enterprise environments, user behavior often shifts due to benign reasons such as team transfers or new access roles. Our system must balance sensitivity to anomalous logic with tolerance to legitimate variance.

    \item \textbf{Control Integration:} Operationalizing detection requires meaningful control responses. Flagging an anomaly is only useful if accompanied by alerting, remediation, or audit hooks.

    \item \textbf{Consensus Drift and Logical Updates:} Over time, even valid behavior may cause opinion drift. Securely updating and sharing \( C_i \) matrices across systems introduces challenges in recalibration, versioning, and privacy-aware synchronization.
\end{itemize}

\subsection{Conclusion.}
This work presents a novel fusion of graph-based opinion dynamics and Bayesian inference for anomaly detection in user-directory systems. By using logical matrices to model per-topic influence and grounding belief evolution in provable convergence theorems, we create a detection signal that is both structurally and statistically aware. The introduction of scaled variance as a volatility signal, combined with Bayesian updating (both static and online), allows us to flag anomalous behavior from deviations in consensus. This bridges the gap between formal multi-agent dynamics and applied behavior monitoring.

\subsection{Next Steps.}
Future efforts will focus on expanding the robustness and deployability of this approach:
\begin{itemize}
    \item \textbf{Enterprise-Scale Simulation:} Scale the system to simulate 1,000+ users and directories, and measure detection latency, convergence behavior, and scoring variance.
    \item \textbf{Semantic Enrichment:} Integrate organizational context (e.g., roles, groups, sensitivity levels) into logic construction and topic similarity.
    \item \textbf{Hybrid Detection:} Combine logic-based scoring with graph embeddings or time-series analysis to reduce false positives and improve interpretability.
    \item \textbf{Explainability:} Develop traceability tools to identify which users, logic dependencies, or directory interactions contributed most to an anomaly score.
    \item \textbf{Security Integration:} Package the framework into a plug-in module for UEBA systems or SIEM pipelines, with hooks for alerting, policy enforcement, and analyst triage.
\end{itemize}

\bibliographystyle{unsrt}
\bibliography{Reference}
\newpage
\begin{algorithm}[tb]
\caption{Dependency-Aware Simulation of Opinion Dynamics}
\label{alg:dependency-sim-1}
\begin{algorithmic}[1]
\State \textbf{Input:} Consensus frame \texttt{cdf}, logic matrices $\{C_i\}$, topic similarity matrix $W$, time steps $T$
\State \textbf{Output:} Final consensus values and opinion trajectories

\State Convert \texttt{cdf} to \texttt{cdfdct}
\State Initialize \texttt{pending\_J\_blocks}, \texttt{completed\_J\_blocks}, \texttt{external\_consensus\_values}
\State Initialize \texttt{results}, \texttt{results\_simple}, set \texttt{iteration} $\gets 0$, \texttt{max\_iters} $\gets 20$

\While{\texttt{pending\_J\_blocks} not empty and \texttt{iteration} $<$ \texttt{max\_iters}}
    \State \texttt{iteration} $\gets$ \texttt{iteration} $+ 1$
    \State Identify \texttt{ready\_J} blocks with dependencies satisfied
    \If{\texttt{ready\_J} is empty}
        \State \textbf{raise} deadlock error
    \EndIf

    \ForAll{$J \in$ \texttt{ready\_J}}
        \State Extract \texttt{topics}, theorem type, and external dependencies
        \State Slice $C_i$ and $W$ accordingly

        \If{Theorem = \texttt{Theorem 2}}
            \State Run \texttt{simulate\_theorem2\_multitopic}
        \ElsIf{Theorem = \texttt{Theorem 3}}
            \State Run \texttt{simulate\_opinion\_dynamics\_singleton}
        \ElsIf{Theorem = \texttt{Theorem 3 (via Corollary 2.1)}}
            \State Gather external consensus values
            \State Run \texttt{simulate\_opinion\_dynamics\_corollary2}
        \ElsIf{Theorem = \texttt{Theorem 4} or external deps exist}
            \State Gather external consensus values
            \State Run \texttt{simulate\_theorem4\_multitopic}
        \Else
            \State \texttt{x\_final}, \texttt{x\_hist} $\gets$ \texttt{None}
        \EndIf

        \ForAll{topic $t$ in \texttt{topics}}
            \State Extract consensus from \texttt{x\_final}
            \If{nearly equal across agents}
                \State Store scalar in \texttt{external\_consensus\_values}
            \Else
                \State Store full opinion vector
            \EndIf
        \EndFor

        \State Record outputs in \texttt{results}, \texttt{results\_simple}
        \State Remove $J$ from \texttt{pending\_J\_blocks}
    \EndFor
\EndWhile

\If{\texttt{iteration} $=$ \texttt{max\_iters}}
    \State Emit early termination warning
\EndIf

\State \Return \texttt{results}, \texttt{results\_simple}
\end{algorithmic}
\end{algorithm}

\end{document}